\documentclass[11pt,letterpaper]{article}
\usepackage{emnlp2017}
\usepackage{times}
\usepackage{latexsym}

\usepackage{tabularx}
\usepackage{booktabs}
\usepackage{graphicx}
\usepackage{amsmath}
\usepackage{tikz}
\usepackage{longtable}
\usepackage{soul}

\makeatletter
\DeclareRobustCommand{\textsupsub}[2]{{%
		\m@th\ensuremath{%
			^{\mbox{\fontsize\sf@size\z@#2}}%
			_{\mbox{\fontsize\sf@size\z@#1}}%
		}%
}}
\makeatother
\newcommand{\result}[3]{\vspace{0.1mm}#1{\small\hspace{1mm}\textsupsub{#2}{#3}}}

\newcolumntype{I}[1]{>{\raggedright\let\newline\\\arraybackslash\hspace{0pt}}p{#1}}
\newcolumntype{R}[1]{>{\raggedleft\let\newline\\\arraybackslash\hspace{0pt}}p{#1}}

\emnlpfinalcopy

\def\emnlppaperid{3}

\title{Encoding Word Confusion Networks with Recurrent Neural Networks\\ for Dialog State Tracking}

\author{Glorianna Jagfeld \and Ngoc Thang Vu \\
	Institute for Natural Language Processing (IMS)\\
	Universit\"at Stuttgart\\
	Pfaffenwaldring 5B\\
	70569 Stuttgart\\
  {\tt \{glorianna.jagfeld,thang.vu\}@ims.uni-stuttgart.de}}

\date{}

\begin{document}

\maketitle

\begin{abstract}
	This paper presents our novel method to encode word confusion networks, which can represent a rich hypothesis space of automatic speech recognition systems, via recurrent neural networks.
	We demonstrate the utility of our approach for the task of dialog state tracking in spoken dialog systems that relies on automatic speech recognition output.
	Encoding confusion networks outperforms encoding the best hypothesis of the automatic speech recognition in a neural system for dialog state tracking on the well-known second Dialog State Tracking Challenge dataset.
	
\end{abstract}

\section{Introduction}

Spoken dialog systems (SDSs) allow users to naturally interact with machines through speech and are nowadays an important research direction, especially with the great success of automatic speech recognition (ASR) systems 
~\cite{ASR_Mohamed2012,ASR_Wayne16}.
SDSs can be designed for generic purposes, e.g. smalltalk~\cite{Eliza_Weizenbaum66,NeuralConvModel_Vinyals15}) or a specific task such as finding restaurants or booking flights~\cite{GUS_Bobrow77,NetworkBasedEnd2EndTaskOriented_Wen16}.
Here, we focus on task-oriented dialog systems, which assist the users to reach a certain goal.

Task-oriented dialog systems are often implemented in a modular architecture to break up the complex task of conducting dialogs into more manageable subtasks.
\citet{DSTCReview_Williams16} describe the following prototypical set-up of such a modular architecture:
First, an ASR system converts the spoken user utterance into text.
Then, a spoken language understanding (SLU) module extracts the user's intent and coarse-grained semantic information.
Next, a dialog state tracking (DST) component maintains a distribution over the state of the dialog, updating it in every turn.
Given this information, the dialog policy manager decides on the next action of the system.
Finally, a natural language generation (NLG) module forms the system reply that is converted into an audio signal via a text-to-speech synthesizer.

Error propagation poses a major problem in modular architectures as later components depend on the output of the previous steps.
We show in this paper that DST suffers from ASR errors, which was also noted by~\citet{NeuralBeliefTracking_Mrksic16}.
One solution is to avoid modularity and instead perform joint reasoning over several subtasks, e.g. many DST systems directly operate on ASR output and do not rely on a separate SLU module~\citep{WordBasedDST_Henderson2014,NeuralBeliefTracking_Mrksic16,DSTMemN2N_Perez17}.
End-to-end systems that can be directly trained on dialogs without intermediate annotations have been proposed for open-domain dialog systems~\cite{NeuralConvModel_Vinyals15}.
This is more difficult to realize for task-oriented systems as they often require domain knowledge and external databases.
First steps into this direction were taken by~\citet{NetworkBasedEnd2EndTaskOriented_Wen16} and ~\citet{TowardsE2EDialogDeepReinforc_ZhaoE16}, yet these approaches do not integrate ASR into the joint reasoning process.

We take a first step towards integrating ASR in an end-to-end SDS by passing on a richer hypothesis space to subsequent components.
Specifically, we investigate how the richer ASR hypothesis space can improve DST.
We focus on these two components because they are at the beginning of the processing pipeline and provide vital information for the subsequent SDS components.
Typically, ASR systems output the best hypothesis or an n-best list, which the majority of DST approaches so far uses~\cite{DST2SoA_Williams14,WordBasedDST_Henderson2014,NeuralBeliefTracking_Mrksic16,LecTrac2_ZilkaJ15}.
However, n-best lists can only represent a very limited amount of hypotheses. 
Internally, the ASR system maintains a rich hypothesis space in the form of speech lattices or confusion networks (cnets)\footnote{\citet{CNet_Mangu00} show that every speech lattice can be converted to a 
cnet without losing relevant hypotheses.  
}.

We adapt recently proposed algorithms to encode lattices with recurrent neural networks (RNNs)~\citep{LatticeRNN_Amazon_Ladhak16,LatticeRNN_NMT_Su16} to encode cnets via an RNN based on Gated Recurrent ~Units (GRUs) to perform DST in a neural encoder-classifier system 
and show that this outperforms encoding only the best ASR hypothesis.
We are aware of two DST approaches that incorporate posterior word-probabilities from cnets in addition to features derived from the n-best lists~\citep{DST2SoA_Williams14,HybridDST_Vodolan17}, but
to the best of our knowledge, we develop the first DST system directly operating on cnets.

\section{Proposed Model}
\begin{figure}[htb]
	\begin{center}
		\includegraphics[width=\linewidth]{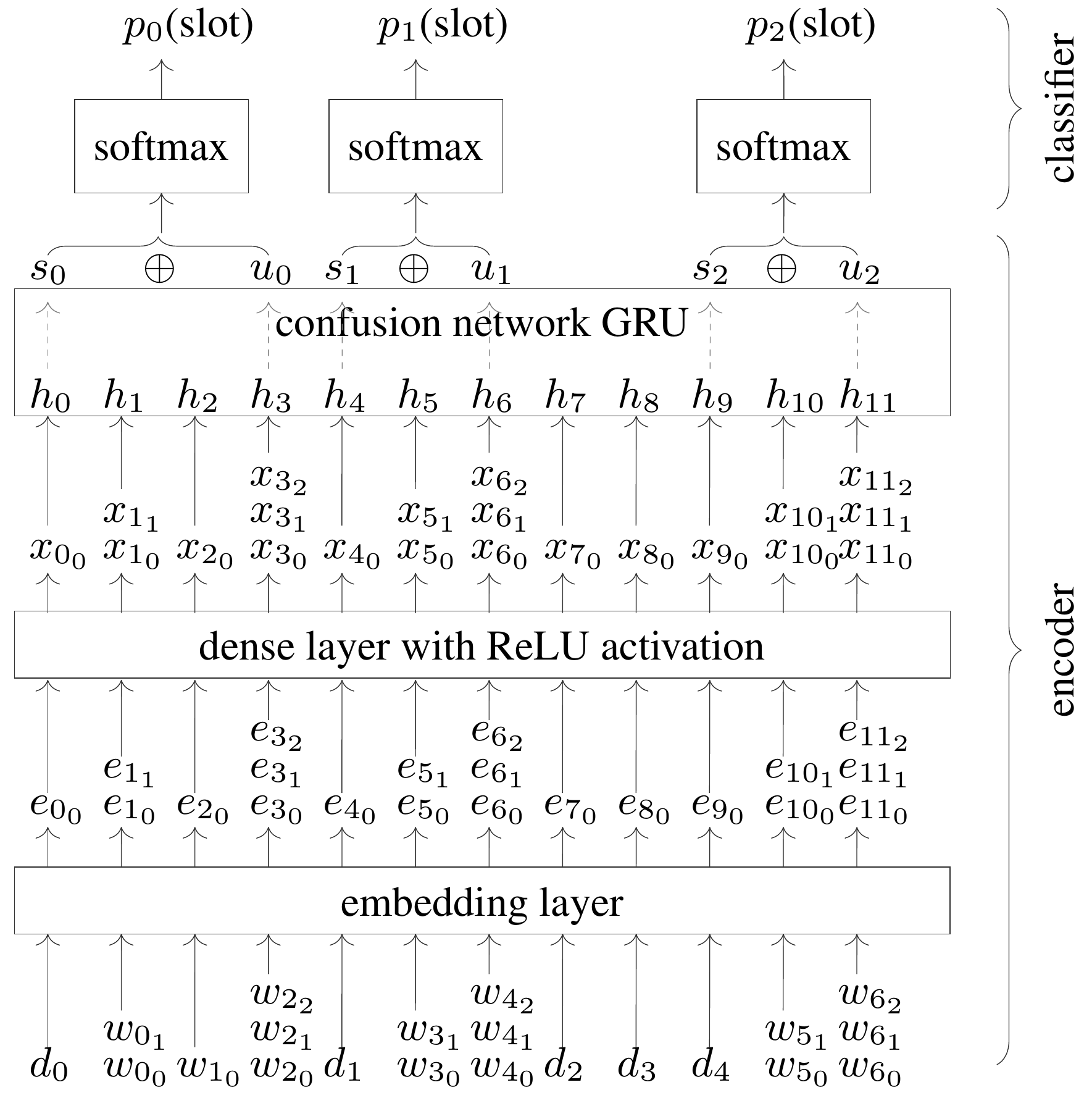}
	\end{center}
	\caption{The proposed model with GRU-based cnet encoder for a dialog with three turns. 
		$d_t$~are one-hot word vectors of the system dialog acts; 
		$w_{t_i}$~correspond to the 
		word hypotheses in the timesteps of the cnets of the user utterances;
		$s_j, u_j$~are the cnet~GRU outputs at the end of each system or user utterance.}
	\label{fig:model}
\end{figure}

Our model depicted in Figure~\ref{fig:model} is based on an incremental DST system~\citep{LecTrac2_ZilkaJ15}.
It consists of an embedding layer for the words in the system and user utterances, followed by a fully connected layer composed of Rectified Linear Units (ReLUs)~\cite{ReLU_Glorot11}, which yields the input to a recurrent layer to encode the system and user outputs in each turn with a softmax classifier on top.
$\oplus$ denotes a weighted sum $c_j$ of the system dialog act $s_j$ and the user utterance $u_j$, 
where $W_s, W_u$, and $b$ are learned parameters:
\begin{equation}
c_j = W_s s_j + W_u  u_j + b
\end{equation}

Independent experiments with the 1-best ASR output showed that a weighted sum of the system and user vector outperformed taking only the user vector $u_j$ as in the original model of~\citet{LecTrac2_ZilkaJ15}.
We chose this architecture over other successful DST approaches that operate on the turn-level of the dialogs~\citep{WordBasedDST_Henderson2014,NeuralBeliefTracking_Mrksic16} because it processes the system and user utterances word-by-word, which makes it easy to replace the recurrent layer of the original version with the cnet encoder.

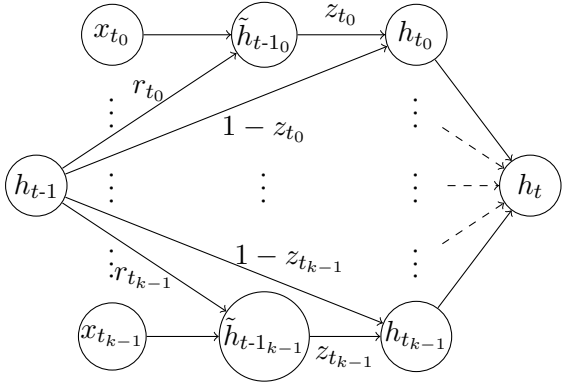
\begin{figure}[htb]
	\begin{tikzpicture}[->] 
	\tikzstyle{neuron}=[circle,draw,minimum size=23pt,inner sep=0pt]
	\tikzstyle{dots}=[circle,minimum size=23pt,inner sep=0pt]
	
	\node[neuron] (hpre) at (0,3) {$h_{t\text{-}1}$};
	
	\node[neuron] (h) at (6.5,3) {$h_t$};
	
	\node[neuron] (xt0) at (1,5) {$x_{t_0}$};
	\node[neuron] (hpret0) at (3,5) {$\tilde{h}_{t\text{-}1_0}$};
	\node[neuron] (ht0) at (5,5) {$h_{t_0}$};
	
	\draw (xt0) -> (hpret0);
	\draw (hpret0) -> (ht0) node[midway, above] {$z_{t_0}$};
	\draw (hpre) -> (hpret0) node[midway, above] {$r_{t_0}$};
	\draw (hpre) -> (ht0) node at (3,3.75) {$1-z_{t_0}$};
	\draw (ht0) -> (h);
	
	\node[neuron] (xtk) at (1,1) {$x_{t_{k-1}}$};
	\node[neuron] (hpretk) at (3,1) {$\tilde{h}_{t\text{-}1_{k-1}}$};
	\node[neuron] (htk) at (5,1) {$h_{t_{k-1}}$};

	\draw (xtk) -> (hpretk);
	\draw (hpretk) -> (htk) node[midway, below] {$z_{t_{k-1}}$};
	\draw (hpre) -> (hpretk) node[midway, right, below] {$r_{t_{k-1}}$};
	\draw (hpre) -> (htk) node[midway, right] {$1-z_{t_{k-1}}$};
	\draw (htk) -> (h);
	
	\node[dots] (xdots1) at (1,4) {\rotatebox{+90}{$\cdots$}};
	\node[dots] (xdots2) at (1,3) {\rotatebox{+90}{$\cdots$}};
	\node[dots] (xdots3) at (1,2) {\rotatebox{+90}{$\cdots$}};
	
	\node[dots] (hpredots2) at (3,3) {\rotatebox{+90}{$\cdots$}};
	
	\node[dots] (hdots1) at (5,4) {\rotatebox{+90}{$\cdots$}};
	\node[dots] (hdots2) at (5,3) {\rotatebox{+90}{$\cdots$}};
	\node[dots] (hdots3) at (5,2) {\rotatebox{+90}{$\cdots$}};
	
	\draw[dashed] (hdots1) -> (h);
	\draw[dashed] (hdots2) -> (h);
	\draw[dashed] (hdots3) -> (h);


%
	
\end{tikzpicture}
	\caption{Encoding $k$~alternative hypotheses at timestep $t$~of a cnet.}
	\label{fig:gru-lattice-deep}
\end{figure}

Our cnet encoder is inspired from two recently proposed algorithms to encode lattices with an RNN with standard memory~\citep{LatticeRNN_Amazon_Ladhak16} and a GRU-based RNN~\citep{LatticeRNN_NMT_Su16}.
In contrast to lattices, every cnet state has only one predecessor and groups together the alternative word hypotheses of a fixed time interval (timestep).
Therefore, our cnet encoder is conceptually simpler and easier to implement than the lattice encoders:
The recurrent memory only needs to retain the hidden state of the previous timestep, while in the lattice encoder the hidden states of all previously processed lattice states must be kept in memory throughout the encoding process.
Following~\citet{LatticeRNN_NMT_Su16}, we use GRUs as they provide an extended memory compared to plain RNNs\footnote{Apart from GRUs, long short-term memory (LSTM) cells~\citep{HochreiterLSTM_97} are a more traditional way to extend the recurrent memory. It is still debated which recurrent memory architecture performs best.
	GRUs are conceptually simpler and have been shown to outperform GRUs for speech signal sequence processing~\citep{ChungGRU_14} and for language modeling with recurrent layers smaller than 200 units~\citep{GRUvsLSTMHighway_Irie16}.
	As our training data is limited, we train models with smaller recurrent layers and therefore use GRUs.
	Yet, we note that the cnet encoding method can be realized with LSTM cells analogously.}.
The cnet encoder reads in one timestep at a time as depicted in Figure~\ref{fig:gru-lattice-deep}.
The key idea is to separately process each of the $k$ word hypotheses representations~$x_{t_i}$ in a timestep with the standard GRU to obtain $k$ hidden states $h_{t_i}$ as defined in Equation~\eqref{eq:gru-deep-output}-\eqref{eq:gru-deep-reset}\footnote{Throughout the paper $\cdot$ denotes an element-wise product.} where $W_z, U_z, b_z, W_h, U_h, b_h, W_r, U_r$, and $b_r$ are the learned parameters of the GRU update, candidate activation and reset gate.
To get the hidden state $h_t$ of the timestep, the hypothesis-specific hidden states $h_{t_i}$ are combined by a pooling function 
(Equation~\ref{eq:gru-deep-pool}).


\begin{align}
h_{t_i} &= z_{t_i} \cdot h_{t-1} + (1-z_{t_i}) \cdot \tilde h_{t_i} 				\label{eq:gru-deep-output}\\
z_{t_i} &= \sigma(W_zx_{t_i} + U_zh_{t-1} + b_z) 					\label{eq:gru-deep-update}\\ 
\tilde h_{t_i} &= \tanh (W_hx_{t_i} + U_h(r_{t_i} \cdot h_{t-1}) + b_h)	\label{eq:gru-deep-cand-output}\\
r_{t_i} &= \sigma(W_rx_{t_i} + U_rh_{t-1} + b_r) 	\label{eq:gru-deep-reset}\\
h_t &= f_{\text{pool}} (h_{t_0} \dots h_{t_{k-1}})\label{eq:gru-deep-pool}
\end{align}

	We experiment with the two different pooling functions $f_{{pool}}$ 
	for the $k$~hidden GRU states~$h_{t_i}$ of the alternative word hypotheses that were used by~\citet{LatticeRNN_Amazon_Ladhak16}:
	
\begin{description}
	\item[average pooling~~~] $f_{{average}} = \frac{\sum_{i=1}^{k} h_{t_i}}{k}$
	\item[weighted pooling~] $f_{{weighted}} = \sum_{i=1}^{k} \text{score}_i \cdot h_{t_i}$, where $\text{score}_i$ is the confidence score of $x_{t_i}$.
\end{description}

Instead of the system output in sentence form we use the dialog act representations in the form of $\langle$dialog-act, slot, value$\rangle$ triples, e.g. \lq inform food Thai\rq, which contain the same information in a more compact way.
Following~\citet{NeuralBeliefTracking_Mrksic16}, we initialize the word embeddings with 300-dimensional semantically specialized PARAGRAM-SL999 embeddings~\cite{ParagramSL999_Wieting15}.
The hyper-parameters for our model are listed in the appendix.

The cnet GRU subsumes a standard GRU-based RNN if each token in the input is represented as a timestep with a single hypothesis.
We adopt this method for the system dialog acts and the baseline model that encode only the best ASR hypothesis.

\section{Data}
In our experiments, we use the dataset provided for the second Dialog State Tracking Challenge (DSTC2)~\cite{DST2_Henderson2014} that consists of user interactions with an SDS in the restaurant domain.
It encompasses 1612, 506, 1117 dialogs for training, development and testing, respectively.
Every dialog turn is annotated with its dialog state encompassing the three goals for \textit{area} (7~values), \textit{food} (93~values) and \textit{price range} (5~values)
and 8~requestable slots, e.g. phone and address.
We train on the manual transcripts and the cnets provided 
with the dataset and evaluate on the cnets.

Some system dialog acts in the DSTC2 dataset do not correspond to words and thus were not included in the pretrained word embeddings.
Therefore, we manually constructed a mapping of dialog acts to words contained in the embeddings, where necessary, e.g. we mapped \textit{expl-conf} to \textit{explicit confirm}.

In order to estimate the potential of improving DST by cnets, we investigated the coverage of words from the manual transcripts for different ASR output types.
As shown in Table~\ref{tab:ASRoutputCoverage}, cnets improve the coverage of words from the transcripts by more than 15~percentage points over the best hypothesis and more than five percentage points over the 10-best hypotheses.

However, the cnets provided with the DSTC2 dataset are quite large.
The average cnet consists of 23~timesteps with 5.5~hypotheses each, amounting to about 125~tokens, while the average best hypothesis contains four tokens.
Manual inspection of the cnets revealed that they contain a lot of noise such as interjections (\textit{uh, oh, ...}) that never appear in the 10-best lists.
The appendix provides an exemplary cnet for illustration.
To reduce the processing time and amount of noisy hypotheses, we remove all interjections and additionally experiment with pruning hypotheses with a score below a certain threshold.
As shown in Table~\ref{tab:ASRoutputCoverage}, this does not discard too many correct hypotheses but markedly reduces the size of the cnet to an average of seven timesteps with two hypotheses.

\begin{table}[htb]
	\begin{center}
		\resizebox{\linewidth}{!}{
			\def \cellwidth {1cm}
			\begin{tabular}{lp{10mm}p{12mm}p{7mm}p{18mm}}
				
				& 1-best & 10-best & cnet & pruned cnet\\
				\midrule
				all words		& 69.3 & 78.6 & 85.7 & 83.1\\
				slots/values 	& 69.8 & 75.6 & 82.4 & 80.6\\
			\end{tabular}
		}
	\end{center}
	\caption{Coverage of words from the manual transcripts in the DSTC2 development set of different \textit{batch}~ASR output types~(\%). 
		In the pruned cnet interjections and hypotheses with scores below 0.001 were removed.}
	\label{tab:ASRoutputCoverage}
\end{table}

\section{Results and Discussion}
We report the joint goals and requests accuracy (all goals or requests are correct in a turn) according to the DSTC2 featured metric~\cite{DST2_Henderson2014}.
We train each configuration 10~times with different random seeds and report the average, minimum and maximum accuracy.
To study the impact of ASR errors on DST, we trained and evaluated our model on the different user utterance representations provided in the DSTC2 dataset.
Our baseline model uses the best hypothesis of the \textit{batch}~ASR system, which has a word error rate (WER) of~34\% on the DSTC2 test set.
Most DST approaches use the hypotheses of the \textit{live}~ASR system, which has a lower WER of~29\%.
We train our baseline on the \textit{batch}~ASR outputs as the cnets were also produced by this system.
As can be seen from Table~\ref{tab:results-dstc2-asr-errors}, the DST accuracy slightly increases for the higher-quality \textit{live}~ASR outputs. 
More importantly, the DST performance drastically increases, when we evaluate on the manual transcripts that reflect the true user utterances nearly perfectly.

	\begin{table}[htb]
	\begin{center}
		\begin{tabular}{lcc}
			\textbf{test data} & \textbf{goals} & \textbf{requests}\\
			\midrule		
			\multicolumn{3}{c}{\textit{train on transcripts + \textit{batch}~ASR (baseline)}}\\
			\midrule
			\textit{batch}~ASR & \result{63.6}{58.7}{66.6} & \result{96.8}{96.5}{97.1}\\
			\midrule
			\multicolumn{3}{c}{\textit{train on transcripts + \textit{live}~ASR}}\\
			\midrule
			\textit{live}~ASR & \result{63.8}{60.2}{67.0} & \result{97.5}{97.2}{97.7}\\
			transcripts & \result{78.3}{74.3}{82.4} & \result{98.7}{98.0}{99.0}\\
		\end{tabular}
	\end{center}
	\caption{DSTC2 test set accuracy for 1-best ASR outputs of ten runs with different random seeds in the format \result{average}{minimum}{maximum}.}
	\label{tab:results-dstc2-asr-errors}
\end{table}

\subsection{Results of the Model with Cnet Encoder}

Table~\ref{tab:results} displays the results for our model evaluated on cnets for increasingly aggressive pruning levels (discarding only interjections, additionally discarding hypotheses with scores below 0.001 and 0.01, respectively).
As can be seen, using the full cnet except for interjections does not improve over the baseline.
We believe that the share of noisy hypotheses in the DSTC2 cnets is too high for our model to be able to concentrate on the correct hypotheses.
However, when pruning low-probability hypotheses both pooling strategies improve over the baseline.
Yet, average pooling performs worse for the lower pruning threshold, which shows that the model is still affected by noise among the hypotheses.
Conversely, the model can exploit a rich but noisy hypothesis space by weighting the information retained from each hypothesis: 
Weighted pooling performs better for the lower pruning threshold 
of~0.001 with which we obtain the highest result overall, improving the joint goals accuracy by 1.6~percentage points compared to the baseline.
Therefore, we  conclude that is beneficial to use information from all alternatives and not just the highest scoring one but that it is necessary to incorporate the scores of the hypotheses and to prune low-probability hypotheses.
Moreover, we see that an ensemble model that averages the predictions of ten cnet~models trained with different random seeds also outperforms an ensemble of ten baseline models.

	\begin{table}[htb]
	\begin{center}
		\begin{tabular}{lcc}
			\textbf{method} & \textbf{goals} & \textbf{requests}\\
			\midrule
			1-best baseline & \result{63.6}{58.7}{66.6} & \result{96.8}{96.5}{97.1}\\
			\midrule
			\multicolumn{3}{c}{\textit{cnet - no pruning}}\\
			\midrule
			weighted pooling & \result{63.7}{61.6}{65.6} & \result{96.7}{96.3}{97.0}\\
			\midrule
			\multicolumn{3}{c}{\textit{cnet - score threshold 0.001}}\\
			\midrule
			average pooling & \result{63.7}{60.0}{66.4} & \result{96.6}{96.0}{96.8}\\
			weighted pooling & \result{\textbf{65.2}}{59.1}{68.5} & \result{97.0}{96.6}{97.4}\\
			\midrule
			\multicolumn{3}{c}{\textit{cnet - score threshold 0.01}}\\
			\midrule
			average pooling & \result{64.6}{59.7}{67.9} & \result{96.9\hphantom{$^\star$}}{96.5}{97.2}\\
			weighted pooling & \result{64.7}{62.2}{68.4} & \result{\textbf{97.1$^\star$}}{96.9}{97.3}\\
			\midrule
			\multicolumn{3}{c}{\textit{ensemble models}}\\
			\midrule
			baseline &   69.7      &      96.7\\
			cnet &  \textbf{71.4} & \textbf{97.2}\\
			\midrule
			\multicolumn{3}{c}{\textit{results from related work}}\\
			\midrule
			\citet{HybridDST_Vodolan17} & 80.0 & -\\ 
			\citet{DST2SoA_Williams14} & 78.4 & 98.0\\ 
			\citet{NeuralBeliefTracking_Mrksic16} & 73.4 & 96.5\\
		\end{tabular}
	\end{center}
	\caption{DSTC2 test set accuracy 
		of ten runs with different random seeds in the format \mbox{\result{average}{minimum}{maximum}}.
		$^\star$~denotes a statistically significant higher result than the baseline ($p<0.05$, Wilcoxon signed-rank test with Bonferroni correction for ten repeated comparisons).
		The cnet ensemble corresponds to the best cnet model with pruning threshold~0.001 and weighted pooling.}
	\label{tab:results}
\end{table}

Although it would be interesting to compare the performance of cnets to full lattices, this is not possible with the original DSTC2 data as there were no lattices provided.
This could be investigated in further experiments by running a new ASR system on the DSTC2 dataset to obtain both lattices and cnets.
However, these results will not be comparable to previous results on this dataset due to the different ASR output.

\subsection{Comparison to the State of the Art}

The current state of the art on the DSTC2 dataset in terms of joint goals accuracy is an ensemble of neural models based on hand-crafted update rules and RNNs~\citep{HybridDST_Vodolan17}.
Besides, this model uses a delexicalization mechanism that replaces mentions of slots or values from the DSTC2 ontology by a placeholder to learn value-independent patterns~\citep{WordBasedDST_Henderson2014,RobustDelexicalisedDST_Henderson2014}.
While this approach is suitable for small domains and languages with a simple morphology
such as English, it becomes increasingly difficult to locate words or phrases corresponding to slots or values in wider domains or languages with a rich morphology~\citep{NeuralBeliefTracking_Mrksic16} and we therefore abstained from delexicalization.

The best result for the joint requests was obtained by a ranking model based on hand-crafted features, which relies on separate SLU systems besides ASR~\citep{DST2SoA_Williams14}.
SLU is often cast as sequence labeling problem, where each word in the utterance is annotated with its role in respect to the user's intent~\cite{SLU_Raymond07,SLU_Vu16}, requiring training data with fine-grained word-level annotations in contrast to the turn-level dialog state annotations.
Furthermore, a separate SLU component introduces an additional set of parameters to the SDS that has to be learned.
Therefore, it has been argued to jointly perform SLU and DST in a single system~\cite{WordBasedDST_Henderson2014}, which we follow in this work.

As a more comparable reference for our set-up, we provide the result of the neural DST system of~\citet{NeuralBeliefTracking_Mrksic16} that like our approach does not use outputs of a separate SLU system nor delexicalized features.
Our ensemble models outperform~\citet{NeuralBeliefTracking_Mrksic16} for the joint requests but are a bit worse for the joint goals. 
We stress that our goal was not to reach for the state of the art but show that DST can benefit from encoding cnets.

\section{Conclusion}
As we show in this paper, ASR errors pose a major obstacle to accurate DST in SDSs.
To reduce the error propagation, we suggest to exploit the rich ASR hypothesis space encoded in cnets that contain more correct hypotheses than conventionally used n-best lists.
We develop a novel method to encode cnets via a GRU-based RNN and demonstrate that this leads to improved DST performance compared to encoding the best ASR hypothesis on the DSTC2 dataset.

In future experiments, we would like to 
explore further ways to leverage the scores of the hypotheses, for example by incorporating them as an independent feature rather than a direct weight in the model.

\section*{Acknowledgments}
We thank our anonymous reviewers for their helpful feedback.
Our work has been supported by the German Research Foundation (DFG) via a research grant to the project A8 within the Collaborative Research Center (SFB) 732 at the University of Stuttgart.

\bibliography{references/DST,references/lattices,references/NN,references/dialogSystems,references/NNCM}
\bibliographystyle{emnlp_natbib}

\clearpage
\newpage

\section*{A. Hyper-Parameters}
\label{sec:supplemental}

\begin{center}
	\resizebox{\linewidth}{!}{
	\begin{tabular}{I{3.3cm}I{3.7cm}}
		parameter & value\\
		\hline
		training epochs & 20 (requests), 50 (area, price range), 100 (food) \\
		optimizer & Adam \\
		initial learning rate &  0.001\\
		training batch size & 10 dialogs\\
		$\lambda$ of l2 regularization & 0.001\\
		dropout rate & 0.5\\
		embeddings & pretrained 300-dimensional PARAGRAM-SL999 embeddings \\
		\# units GRU & 100\\
		\# units dense layer & 300\\
		size of the system and user vector combination matrix & 50\\
		user utterance type training & transcript + cnet \\
		user utterance type testing & cnet\\
	\end{tabular}
	}
\end{center}

\newpage
\clearpage
\newpage
\begin{table*}[h!]
	\begin{large}\textbf{B. Cnet from the DSTC2 Dataset}\end{large} \\
	\begin{center}
		\resizebox*{!}{0.73\paperheight}{
		\begin{tabular}{lllp{14.5cm}}
			& start & end & hypotheses with scores \\
			\midrule
1&0.0328125&0.0492188&!null (-0.0001) uh (-31.83215) ah (-32.41007) i (-34.84077) oh (-40.73034) a (-41.20651) \\
2&0.0492188&0.065625&!null (-0.0001) i (-36.65728) uh (-48.94583) ah (-52.79816) oh (-55.63619) \\
3&0.065625&0.0820312&!null (-0.0001) oh (-47.15494) \\
4&0.0820312&0.0984375&!null (-0.0001) and (-47.59002) \\
5&0.0984375&0.13125&!null (-0.0001) ah (-33.03135) uh (-39.74279) i'm (-41.90521) i (-42.4907) ok (-42.98212) and (-43.31765) can (-45.37124) \\
6&0.13125&0.1476562&!null (-0.0001) um (-30.17054) i'm (-32.94894) uh (-35.07708) i (-36.82227) can (-36.89635) and (-36.99255) ah (-43.84253) \\
7&0.1476562&0.1640625&!null (-0.0001) ah (-41.90521) \\
8&0.1640625&0.196875&!null (-0.0001) and (-31.41877) ah (-33.03021) i (-34.15576) um (-37.12041) i'm (-37.5037) uh (-40.89799) can (-42.66815) \\
9&0.196875&0.2296875&!null (-0.0001) ok (-37.41767) i (-43.27491) \\
10&0.2296875&0.2625&!null (-0.0001) uh (-28.98055) and (-30.48886) i (-30.50464) ah (-31.02539) can (-31.49024) a (-31.74998) um (-39.56715) i'm (-39.6478) \\
11&0.2625&0.2707031&!null (-0.0001) a (-48.38457) \\
12&0.2707031&0.2789062&!null (-0.0001) i (-45.51492) \\
13&0.2789062&0.2953125&!null (-0.0001) uh (-37.77175) \\
14&0.2953125&0.328125&!null (-0.0001) uh (-22.47343) and (-24.25971) i (-25.13368) can (-31.76437) um (-32.11736) oh (-32.22958) is (-32.77696) ah (-36.18502) \\
15&0.328125&0.3445312&!null (-0.0001) ah (-25.74752) uh (-29.74647) i (-35.53291) um (-37.89059) oh (-40.87821) \\
16&0.3445312&0.3609375&!null (-0.0001) uh (-21.97038) oh (-31.83063) ah (-31.96235) i (-42.61901) \\
17&0.3609375&0.39375&!null (-0.0001) ah (-24.38169) and (-24.39148) ok (-25.08438) i (-29.82585) can (-30.21743) i'm (-33.53017) \\
18&0.39375&0.525&!null (-0.0001) uh (-23.14362) i (-24.16806) can (-24.21132) um (-24.52006) it (-29.71162) ok (-31.79314) ah (-33.52439) and (-36.14101) \\
19&0.525&0.590625&!null (-0.0001) ah (-52.30994) \\
20&0.590625&0.65625&!null (-0.0001) uh (-26.81306) \\
21&0.65625&0.7875&!null (-0.0001) uh (-17.00693) can (-18.18777) i (-21.7525) and (-22.92453) a (-23.86453) in (-26.00351) ok (-32.25924) ah (-33.28463) it (-37.21361) oh (-45.34864) \\
22&0.7875&0.8039062&!null (-0.0001) i (-18.35259) and (-18.3801) a (-19.56405) it (-20.65148) is (-20.78921) uh (-22.80336) ok (-23.32806) can (-24.81112) oh (-28.52324) \\
23&0.8039062&0.8203125&!null (-0.0001) i (-32.22319) \\
24&0.8203125&0.853125&!null (-0.0001) uh (-9.748239) i (-12.90367) ah (-15.49612) ok (-15.62111) can (-19.96378) and (-23.52033) \\
25&0.853125&0.8859375&!null (-0.0001) and (-10.25172) uh (-10.51098) i (-14.77064) ok (-17.1938) it (-17.42765) ah (-24.74307) \\
26&0.8859375&0.91875&!null (-0.0001) ok (-10.7207) and (-14.63778) i (-17.40079) \\
27&0.91875&0.984375&!null (-0.005078796) and (-5.305283) ok (-9.687913) can (-10.20153) is (-13.44094) uh (-17.34175) where (-23.62194) \\
28&0.984375&1.05&!null (-0.009671085) ok (-5.591656) could (-5.726142) can (-5.96063) and (-9.760586) it (-17.42122) \\
29&1.05&1.13&i (-0.003736897) !null (-5.591568) i'd (-14.10718) ok (-20.44036) could (-21.03084) \\
30&1.13&1.21&!null (-0.003736222) i (-5.59171) could (-15.09615) i'd (-15.67228) thank (-16.10791) it (-16.47987) \\
31&1.21&1.34&don't (-0.0001) !null (-14.78975) know (-24.44728) gone (-27.63221) i (-28.97229) a (-32.95747) go (-41.58155) da (-47.35928) \\
32&1.34&1.405&!null (-0.0001) don't (-14.78604) i (-23.63712) a (-24.3221) are (-25.11523) it (-27.08631) uh (-31.06854) of (-32.07071) \\
33&1.405&1.4375&!null (-0.0001) of (-17.31417) a (-22.29353) ok (-25.30747) i (-30.73294) are (-31.25772) \\
34&1.4375&1.47&!null (-0.0001) tv (-24.90913) a (-31.64189) \\
35&1.47&1.5975&care (-0.0001) t (-13.25217) i (-16.79167) to (-19.88062) !null (-22.45499) \\
36&1.5975&1.725&!null (-0.0001) care (-15.73215) \\
37&1.725&1.78875&!null (-0.002474642) for (-6.446757) of (-7.396389) food (-8.225521) care (-12.98698) if (-13.04223) and (-16.05245) i (-16.57308) kind (-16.92007) uh (-17.26407) a (-18.45659) or (-18.46813) are (-18.88889) tv (-27.09801) \\
38&1.78875&1.8525&!null (-0.0001) i (-13.25853) in (-14.35854) of (-17.30617) uh (-20.08914) and (-20.30067) tv (-21.15766) a (-25.55673) \\
39&1.8525&1.91625&!null (-0.0004876809) the (-7.78335) food (-9.733769) for (-11.98406) i (-12.23129) i'm (-14.38366) of (-18.23437) and (-19.87061) \\
40&1.91625&1.98&!null (-0.0001) of (-11.92066) the (-11.98383) food (-12.77184) for (-14.38366) \\
		\end{tabular}
	}
	\end{center}
	\caption{Cnet from the DSTC2 development set of the session with id voip-db80a9e6df-20130328\_230354.
		The transcript is \textit{i don't care}, which corresponds the best hypothesis of both ASR systems.
		Every timestep contains the hypothesis that there  is no word (!null).}
	\label{fig:cnet}
\end{table*}

\end{document}